\pdfoutput=1

\documentclass[11pt]{article}

\usepackage[]{ACL2023}

\usepackage{times}
\usepackage{latexsym}

\usepackage[T1]{fontenc}

\usepackage[utf8]{inputenc}

\usepackage{microtype}

\usepackage{multirow}
\usepackage{booktabs}
\usepackage{amsfonts}
\usepackage{subfigure}
\usepackage{graphicx}
\usepackage{tabularx}
\usepackage{amssymb}
\usepackage{amsmath}
\usepackage{amsthm}
\usepackage{bm}
\usepackage{stmaryrd}

\usepackage{algorithm}
\usepackage{algorithmic}
\usepackage{arydshln}
\usepackage{threeparttable}
\usepackage{epstopdf}
\usepackage{bigstrut,multirow,rotating}
\usepackage{color}
\usepackage{makecell}

\usepackage{inconsolata}

\title{Towards Adaptive Prefix Tuning\\
for Parameter-Efficient Language Model Fine-tuning}

\author{
Zhen-Ru Zhang,
Chuanqi Tan,
Haiyang Xu,
Chengyu Wang,
Jun Huang,
Songfang Huang \\
Alibaba Group\\
\texttt{\{zhangzhenru.zzr,chuanqi.tcq,shuofeng.xhy\}@alibaba-inc.com}\\
\texttt{\{chengyu.wcy,huangjun.hj,songfang.hsf\}@alibaba-inc.com}
}

\begin{document}
\maketitle
\begin{abstract}
Fine-tuning large pre-trained language models on various downstream tasks with whole parameters is prohibitively expensive. Hence, Parameter-efficient fine-tuning has attracted attention that only optimizes a few task-specific parameters with the frozen pre-trained model. In this work, we focus on prefix tuning, which only optimizes continuous prefix vectors (i.e. pseudo tokens) inserted into Transformer layers. Based on the observation that the learned syntax and semantics representation varies a lot at different layers, we argue that the adaptive prefix will be further tailored to each layer than the fixed one, enabling the fine-tuning more effective and efficient. Thus, we propose Adaptive Prefix Tuning (APT) to adjust the prefix in terms of both fine-grained token level and coarse-grained layer level with a gate mechanism. Experiments on the SuperGLUE and NER datasets show the effectiveness of APT. In addition, taking the gate as a probing, we validate the efficiency and effectiveness of the variable prefix.
\end{abstract}

\section{Introduction}\label{intro}
Vanilla fine-tuning strategy usually adjusts all the parameters to adapt the pre-trained language model to downstream tasks. Parameter-efficient learning \cite{he2022towards, pmlr-v97-houlsby19a, lester-etal-2021-power, guo-etal-2021-parameter, ben-zaken-etal-2022-bitfit} is an emerging framework that freezes the pre-trained model and only tunes a few number of task-specific parameters for downstream tasks. For instance, Prefix tuning \cite{li-liang-2021-prefix, liu-etal-2022-p} prepends length-equivalent pseudo prefix tokens, i.e. continuous task-specific vectors to each layer of the pre-trained model, achieving comparable even superior performance with only 0.1-3\% parameters.

\begin{figure}[t]
    \centering
    \includegraphics[width=7.7cm]{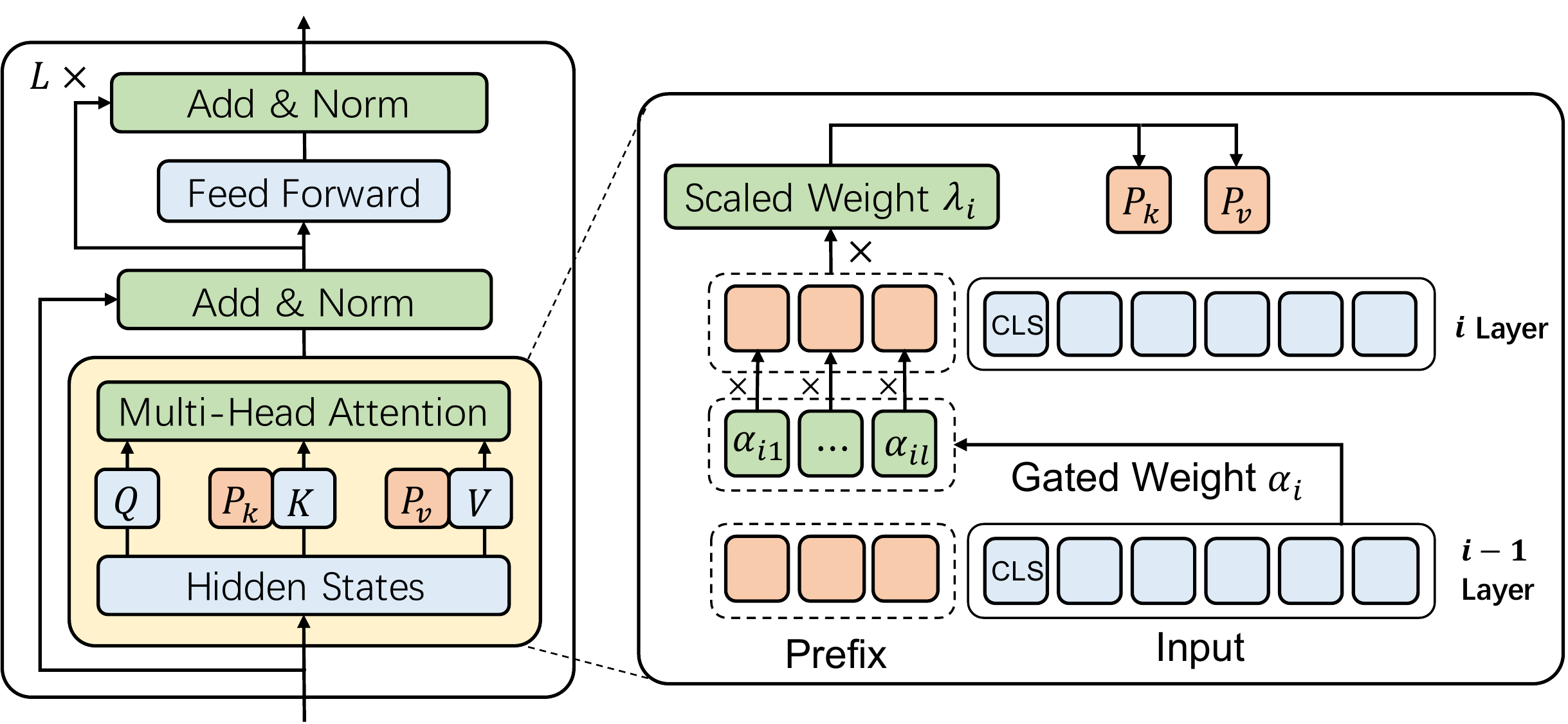}
    \caption{An illustration of the proposed approach APT where the left is the internal structure of Transformer with inserted prefixes, and the right is the schematic of prefix gate mechanism.}
    \label{adaptive prefix}
\end{figure}

In previous works, the length of prefix tokens (or the number of trainable parameters) is usually the same at each layer. However, a potential observation lies in that the structure information and representational capacity embedded in each layer are prone to be inconsistent \cite{jawahar-etal-2019-bert}. It is generally considered that the bottom layers of the language model tend to capture concrete and shallow phrase-level features, while the top layers concerns more with abstract semantic information \cite{, tenney-etal-2019-bert}. Based on the perspective, we assume adaptive prefix can grab the emphasis more flexibly to adapt to various downstream tasks.

In light of above motivation, we investigate the adaptive prefix in this work. We propose \textbf{A}daptive \textbf{P}refix \textbf{T}uning (APT) with an adaptive gate mechanism at both fine-grained token level and coarse-grained layer level. Specifically, as shown in Figure~\ref{adaptive prefix}, for fine granularity, APT scores each individual prefix token via gated weight assignment. Then, the scaled weight is utilized to balance the inserted task-specific prefix tokens and original input tokens for current layer at coarse-grained level.

Extensive experiments against prefix tuning on the sentence and token classification tasks in full data and low resources setting validate the effectiveness of APT. In addition, the gate learned from APT could be served as a probing for the number of necessary parameters in different layers, guiding us to directly apply variable prefix to the original prefix tuning. The probing experiment further demonstrates the effectiveness of adaptive prefix.

\section{Related Works}
Since fine-tuning the whole model is prohibitively expensive, parameter-efficient language model fine-tuning becomes a lightweight alternative that only optimizes a small number of parameters while keeping most pre-trained parameters frozen \cite{he2022towards}. 
Adapter tuning \cite{pmlr-v97-houlsby19a} inserts two tunable task-specific modules after multi-head attention and feed-forward network, achieving comparable performance with only 2-4\% of the parameters. Prompt tuning \cite{lester-etal-2021-power} and Prefix-Tuning \cite{li-liang-2021-prefix} only train soft prompts by adding prefix tokens to the input or hidden states. 
Recently, \citet{liu-etal-2022-p} extend the prefix tuning to the natural language understanding tasks, which matches the performance of fine-tuning with only 0.1\%-3\% tuned parameters. 

Furthermore, with an overlap of our motivations that each layer of the pre-trained language model focuses on different aspects of feature for various tasks \cite{jawahar-etal-2019-bert, clark-etal-2019-bert} and extra parameters are probably not necessary for certain tasks \cite{pmlr-v97-houlsby19a, Fan2020ReducingTD, ruckle-etal-2021-adapterdrop}, Adaptable Adapters \cite{moosavi_adaptable_2022} selects beneficial adapter layers and learns task-specific activation function for downstream tasks to make adaptor dynamic for each task and layer. In addition to different frameworks (adapter versa prefix tuning), our key difference from their work lies in that we aim to dynamically filter required information at each layer in a soft way, while they choose whether to add trainable modules at the layer level in a hard manner.

\begin{table*}[htp]
\centering
\small
\resizebox{1\textwidth}{!}{
\begin{tabular}{@{}lccccccccccccc@{}}
\toprule
\multirow{2}{*}{\textbf{Model}} & \multirow{2}{*}{} & \multicolumn{6}{c}{\textbf{SuperGLUE}} & \multicolumn{4}{c}{\textbf{NER}} \\ \cmidrule(lr){3-8} \cmidrule(lr){9-12} 
 &  & BoolQ & COPA & RTE & WiC & WSC & Avg. & CoNLL03 & CoNLL04 & OntoNotes & Avg. \\ \midrule
\multirow{3}{*}{\thead[l]{BERT-base \\ (110M)}} & FT & \textbf{72.9} & 67.0 & 68.4 & \underline{71.1} & 63.5 & 68.6 & - & - & - & - \\
 & PT-2 & 72.5 & \underline{67.4} & \underline{71.3} & 69.5 & \underline{65.4} & \underline{69.2} & 89.3 & 82.6 & 87.1 & 86.3 \\
 & APT & \underline{72.6} & \textbf{70.0} & \textbf{72.7} & \textbf{71.2} & \textbf{66.9} & \textbf{70.7} & \textbf{89.7} & \textbf{84.1} & \textbf{87.2} & \textbf{87.0} \\ \midrule
\multirow{3}{*}{\thead[l]{BERT-large \\ (335M)}} & FT & \textbf{77.7} & 69.0 & 70.4 & \underline{74.9} & 68.3 & 72.1 & \textbf{92.8} & \underline{85.6} & \textbf{89.2} & \textbf{89.2} \\
 & PT-2 & 75.8 & \underline{73.0} & \underline{78.3} & \textbf{75.1} & 68.3 & \underline{74.1} & 90.2 & 84.5 & 86.4 & 87.0 \\
 & APT & \underline{76.0} & \textbf{79.0} & \textbf{79.4} & \textbf{75.1} & \textbf{70.2} & \textbf{75.9} & \underline{90.7} & \textbf{85.8} & \underline{88.6} & \underline{88.4} \\ \midrule
\multirow{3}{*}{\thead[l]{RoBRETa-large \\ (355M)}} & FT & \textbf{86.9} & \textbf{94.0} & 86.6 & \textbf{75.6} & 63.5 & \underline{81.3} &  92.6 & \underline{88.8} & \textbf{89.8} & \underline{90.4} \\
 & PT-2 & \underline{84.8} & \underline{93.0} & \underline{89.5} & 73.4 & 63.5 & 80.8 & \textbf{92.8} & 88.4 & \textbf{89.8} & 90.3 \\
 & APT & \underline{84.8} & \textbf{94.0} & \textbf{89.9} & \underline{74.6} & \textbf{68.3} & \textbf{82.3} & \underline{92.7} & \textbf{89.0} & \textbf{89.8} & \textbf{90.5} \\ \midrule
\multirow{3}{*}{\thead[l]{DeBERTa-xlarge\\ (750M)}} & FT & - & - & - & - & - & - & \textbf{93.1} & \textbf{89.1} & \underline{90.4} & \textbf{90.9} \\
 & PT-2 & - & - & - & - & - & - & \textbf{93.1} & \underline{86.5} & \underline{90.4} & 90.0 \\
 & APT & - & - & - & - & - & - & \underline{93.0} & \textbf{89.1} & \textbf{90.5} & \underline{90.8} \\ \bottomrule
 
\end{tabular}
}
\caption{The results on SuperGLUE development set and NER test set in full data setting. The metric of SuperGLUE is accuracy and other is micro-f1 score. Results for FT and PT-2 on BERT-large, RoBRETa-large and DeBERTa-xlarge are token from \cite{liu-etal-2022-p}. Results for FT on BERT-base are from \cite{liu2021gpt}. (FT: vanilla fine-tuning; PT-2: P-Tuning v2; APT: Adaptive Prefix Tuning; \textbf{bold}: the best score; \underline{underline}: the second best)}
\label{full data results}
\end{table*}

\begin{table*}[t]
\centering
\small
\begin{tabular}{cccccccc}
\toprule
\textbf{Setting} & \textbf{Method} & BoolQ & COPA & RTE & WiC & WSC & Avg. \\
\midrule
\multirow{3}{*}{\thead{BERT-base \\ (16-shot)}} & FT &47.2$_{7.5}$ &54.0$_{6.5}$&49.4$_{2.7}$&50.3$_{2.3}$&46.2$_{6.8}$ & 49.4 \\
& PT-2 & 52.4$_{7.2}$ & 54.2$_{3.3}$ & 50.8$_{3.1}$ & 48.2$_{3.3}$ & 48.5$_{4.3}$ & 50.8 \\
& APT & \textbf{55.7$_{6.5}$} & \textbf{57.4$_{2.7}$} & \textbf{53.1$_{4.4}$} & \textbf{53.7$_{2.2}$} & \textbf{55.2$_{3.8}$} & \textbf{55.0} \\ \midrule
\multirow{3}{*}{\thead{BERT-large \\ (16-shot)}} & FT & \textbf{57.3$_{9.7}$} & 52.0$_{2.4}$&49.5$_{2.7}$&50.0$_{0.0}$&38.7$_{2.2}$ & 49.5 \\
&  PT-2 & 50.3$_{5.7}$ & 58.2$_{5.3}$ & 49.9$_{3.4}$ & 49.3$_{2.2}$ & 48.1$_{4.2}$ & 51.2 \\
& APT & 51.7$_{3.5}$ & \textbf{60.0$_{6.3}$} & \textbf{53.9$_{4.6}$} & \textbf{51.8$_{4.8}$} & \textbf{55.4$_{2.3}$} & \textbf{54.6} \\ \midrule
\multirow{3}{*}{\thead{BERT-base \\ (32-shot)}} & FT &48.1$_{9.4}$&52.2$_{6.4}$&49.5$_{2.7}$&49.4$_{0.9}$& \textbf{60.4$_{3.8}$} &51.9 \\
&  PT-2 & 50.1$_{5.5}$ & 55.0$_{3.2}$ & 53.8$_{3.4}$ & 52.0$_{4.1}$ & 51.5$_{4.6}$ & 52.5 \\
& APT & \textbf{53.5$_{5.3}$} & \textbf{57.6$_{2.2}$} & \textbf{56.5$_{1.6}$} & \textbf{54.8$_{3.9}$} & 54.6$_{6.5}$ & \textbf{55.4} \\ \midrule
\multirow{3}{*}{\thead{BERT-large \\ (32-shot)}} & FT &47.6$_{11.9}$&45.0$_{3.6}$&48.4$_{2.2}$&50.0$_{0.0}$&47.3$_{13.2}$ & 47.6 \\
& PT-2 & 45.5$_{5.1}$ & 57.4$_{6.9}$ & 51.3$_{2.3}$ & 53.3$_{2.1}$ & 46.0$_{7.1}$ & 50.7 \\
& APT & \textbf{49.9$_{5.9}$} & \textbf{62.0$_{5.0}$} & \textbf{55.5$_{3.6}$} & \textbf{54.9$_{2.8}$} & \textbf{49.0$_{4.4}$} & \textbf{54.3} \\ \bottomrule
\end{tabular}
\caption{The mean$_{std}$ experimental results within 5 random seeds on SuperGLUE development set in 16-shot and 32-shot setting where all metrics are accuracy. \textbf{bold}: the best score.}
\label{low resources results}
\end{table*}

\begin{table*}[t]
\small
\centering
\begin{tabular}{@{}lcccccccccc@{}}
\toprule
\multirow{2}{*}{\textbf{Setting}} & \multicolumn{6}{c}{\textbf{SuperGLUE}} & \multicolumn{4}{c}{\textbf{NER}} \\ \cmidrule(lr){2-7} \cmidrule(lr){8-11}
& BoolQ & COPA & RTE & WiC & WSC & Avg. & CoNLL03 & CoNLL04 & OntoNotes & Avg. \\ \midrule
APT & \textbf{72.6} & \textbf{70.0} & \textbf{72.7} & \textbf{71.2} & \textbf{66.9} & \textbf{70.7} & \textbf{89.7} & \textbf{84.1} & \textbf{87.2} & \textbf{87.0} \\
\midrule
w/o token-level $\bm{\alpha}$ & 72.6 & 69.0 & 69.9 & 70.8 & 65.8 & 69.6 & 89.5 & 83.7 & 87.2 & 86.8 \\
w/o layer-level $\lambda$ & 72.1 & 67.4 & 71.3 & 69.6 & 65.4 & 69.1 & 89.0 & 82.6 & 86.9 & 86.2 \\
w/o hidden states $\bm{h}$ & 72.0 & 68.8 & 68.7 & 70.2 & 64.6 & 68.9 & 89.1 & 83.6 & 87.1 & 86.6 \\
\bottomrule
\end{tabular}
\caption{Ablation study on BERT-base for two different level gate mechanisms and the hidden states from the previous layer. \textbf{bold}: the best score.}
\label{ablation study}
\end{table*}

\begin{table*}[htp]
\centering
\small
\begin{tabular}{@{}lcccccccccc@{}}
\toprule
\multirow{2}{*}{\textbf{Model}} & \multicolumn{6}{c}{\textbf{SuperGLUE}} & \multicolumn{4}{c}{\textbf{NER}} \\ \cmidrule(lr){2-7} \cmidrule(lr){8-11}
 & BoolQ & COPA & RTE & WiC & WSC & Avg. & CoNLL03 & CoNLL04 & OntoNotes & Avg. \\ \midrule
PT-2 & 72.5 & 67.4 & 71.3 & 69.5 & 65.4 & 69.2 & 89.3 & 82.6 & 87.1 & 86.3 \\
PT-2* & \textbf{72.6} & \textbf{68.8} & \textbf{71.9} & \textbf{70.0} & \textbf{65.8} & \textbf{69.8} & \textbf{89.3} & \textbf{83.0} & \textbf{87.2} & \textbf{86.5} \\
\midrule
PT-2+ & \textbf{72.8} & 65.4 & 69.1 & 71.1 & 65.8 & 68.8 & 89.4 & 83.2 & 87.1 & 86.6 \\
APT & 72.6 & \textbf{70.0} & \textbf{72.7} & \textbf{71.2} & \textbf{66.9} & \textbf{70.7} & \textbf{89.7} & \textbf{84.1} & \textbf{87.2} & \textbf{87.0} \\ \bottomrule
\end{tabular}
\caption{Comparison between PT-2 and PT-2$^{*}$, PT-2$^{+}$ and APT on BERT-base. (PT-2: P-Tuning v2; PT-2$^{*}$: PT-2 with variable prefix; PT-2$^{+}$: PT-2 with enlarged prefix)}
\label{discussion results}
\end{table*}

\section{Methodology}
\subsection{Prefix Tuning}
As prefix tuning is an extension on Transformer \cite{NIPS2017_3f5ee243}, we first recap the structure of Transformer. Transformer is the block consisting of multi-head attention concatenated by multiple single self-attention functions and a fully connected feed-forward network. Formally speaking, the Transformer block is calculated as follows:
\begin{align}\label{self-attention}
{\rm Attn}(\bm{Q}, \bm{K}, \bm{V}) = {\rm softmax}(\frac{\bm{QK}^T}{\sqrt{d}}\bm{V}) \\
{\rm FFN}(\bm{x}) = {\rm ReLU}(\bm{x}\bm{W}_1+\bm{b}_1)\bm{W}_2+\bm{b}_2
\end{align}

Prefix tuning prepends pseudo prefix tokens of length $l$ to each layer of the language model, which is implemented by concatenating inserted keys and values matrix with original corresponding items in each multi-head attention. Specifically, let $\bm{P}_k, \bm{P}_v \in \mathbb{R}^{l \times d} $ be the keys and values of the engaged prefix separately, where $l$ denotes the length of prefix and $d$ corresponds to the dimension, thus self-attention function can be reformatted as:
\begin{eqnarray}\label{self-attention for prefix tuning}
{\rm Attn}(\bm{Q}, \bm{K}^{\prime}, \bm{V}^{\prime})={\rm softmax}(\frac{\bm{Q}(\bm{K}^{\prime})^T}{\sqrt{d}}\bm{V}^{\prime}) \\
\nonumber {\rm where}~~\bm{K}^{\prime}=[\bm{P}_k; \bm{K}], \bm{V}^{\prime}=[\bm{P}_v; \bm{V}] \hspace{10pt}
\end{eqnarray}
Here, $[;]$ donates concatenation function.

\subsection{Adaptive Prefix Tuning}
The length of prefix is usually a manually set hyper-parameter for each task and fixed in distinct layers of the model. However, existing work demonstrates each layer of the language model pays attention to different aspects of the input feature. We assume the prefix in fixed length is insufficient to tailor different layers and tasks. To dynamically customize the prefix at each layer, APT performs a gate mechanism via fine-grained gated weight assignment and coarse-grained scaled weight specification.

Specifically, to capture the diversity of information utilization at different layers, we go deep into the token level at the fine-grained granularity. The token-level gate can inspire us on how many trainable parameters (i.e. pseudo tokens in prefix tuning) are required for this layer, which will be discussed in Section \ref{discussion1}. Thus, APT yields the gated weights of $l$ pseudo tokens at each layer. We use the hidden states to represent the information encoded in the layer and calculate the gated weights $\bm{\alpha}_{i}=[\alpha_{i1},\alpha_{i2},\ldots,\alpha_{il}]$ for $i$-th layer as:
\begin{eqnarray}\label{gated weight}
\bm{\alpha}_{i} = {\rm sigmoid}(\bm{h}_{i-1} \bm{W}_{i})
\end{eqnarray}
Here, $\bm{h}_{i-1}$ is the $d$-dimensional hidden states from the previous layer, and $\bm{W}_i \in \mathbb{R}^{d \times l}$ corresponds to the parameters to be learned.

Besides, we also design a coarse-level gate to balance the information brought from task-specific prefix tokens and original input tokens by learning a layer-level weight. A learnable scaled weight $\lambda_i$ is added to the representation of pseudo prefix tokens at the $i$-th layer.

With the above strategy, the keys-values pair $\bm{P}_i=[\bm{P}_{ik}, \bm{P}_{iv}]$ derived from pseudo prefix tokens in $i$-th layer is updated to $\hat{\bm{P}}_{i}$ as:
\begin{eqnarray}\label{adaptive prefix key-value pair}
\hat{\bm{P}}_{i}=\lambda_{i} \bm{\alpha}_{i} \odot [\bm{P}_{ik}, \bm{P}_{iv}]
\end{eqnarray}
$\odot$ is the element-wise multiplication. Accordingly, the calculation of the self-attention function in APT is similar to Eq.(\ref{self-attention for prefix tuning}) without further elaboration.

\section{Experiments}
\subsection{Experimental Setup}
We conduct 5 NLU tasks on SuperGLUE \cite{NEURIPS2019_4496bf24} benchmark including BoolQ \cite{clark-etal-2019-boolq}, COPA \cite{roemmele2011choice}, RTE \cite{wang-etal-2018-glue}, WiC \cite{pilehvar-camacho-collados-2019-wic} and WSC \cite{levesque2012winograd} as well as 3 Named Entity Recognition (NER) tasks including CoNLL03 \cite{tjong-kim-sang-de-meulder-2003-introduction}, CoNLL04 \cite{carreras-marquez-2004-introduction}, and OntoNotes 5.0 \cite{AB2/MKJJ2R_2013}. With BERT-base / large \cite{devlin-etal-2019-bert} and RoBERTa-large \cite{liu2019roberta} instantiated by HuggingFace Transformers \cite{wolf-etal-2020-transformers}, we compare APT with vanilla fine-tuning and P-Tuning v2 \cite{liu-etal-2022-p} which is an implementation of the prefix tuning, configured with hyper-parameters public in the released code\footnote{\url{https://github.com/THUDM/P-tuning-v2}}. We also verify our method with DeBERTa-xlarge \cite{he2020deberta} on NER tasks following P-Tuning v2.

\subsection{Results}
We report the main results in Table \ref{full data results}. For BERT-base, we can observe that APT achieves 1.5\% and 0.7\% improvements over P-Tuning v2 on SuperGLUE and NER tasks, respectively. For BERT-large, APT outperforms P-Tuning v2 by 1.8\% on SuperGLUE tasks and 1.4\% on NER tasks. For RoBERTa-large, APT surpasses P-Tuning v2 by 1.5\% on SuperGLUE tasks and 0.2\% on NER tasks. On NER tasks with DeBERTa-xlarge, APT is superior to P-Tuning v2 by an average of 0.8\%. Compared with vanilla fine-tuning, APT is comparable or even better on part of tasks. In addition, we explore the experimental performance under low resource settings on SuperGLUE benchmark. As shown in Table \ref{low resources results}, APT is a better few-shot learner than P-Tuning v2, which exceeds 4.2\%, 3.4\% in 16-shot setting, and 2.9\%, 3.6\% in 32-shot setting for BERT-base and BERT-large, respectively.

\subsection{Ablation Study}
\label{Ablation Study Section}
We conduct an ablation study in order to explore the separate effect of token-level gated weight $\bm{\alpha}$, layer-level scaled weight $\lambda$ and the hidden states $\bm{h}$ from the previous layer which is used to calculate token-level gated weight $\bm{\alpha}$ in Eq.(\ref{gated weight}). As shown in Table \ref{ablation study}, it can be found that removing any strategy hurts the performance to varying degrees, demonstrating that they are all advantageous. Specifically, the beneficial effect of $\lambda$ for APT is slightly greater than $\bm{\alpha}$ overall. Besides, it is effective and meaningful to introduce the context (i.e. the hidden states $\bm{h}$ from the previous layer) when obtaining the gated weight, especially for SuperGLUE tasks.

\begin{figure}[t]
    \centering
    \subfigure[COPA]{
       \includegraphics[width=0.22 \textwidth]{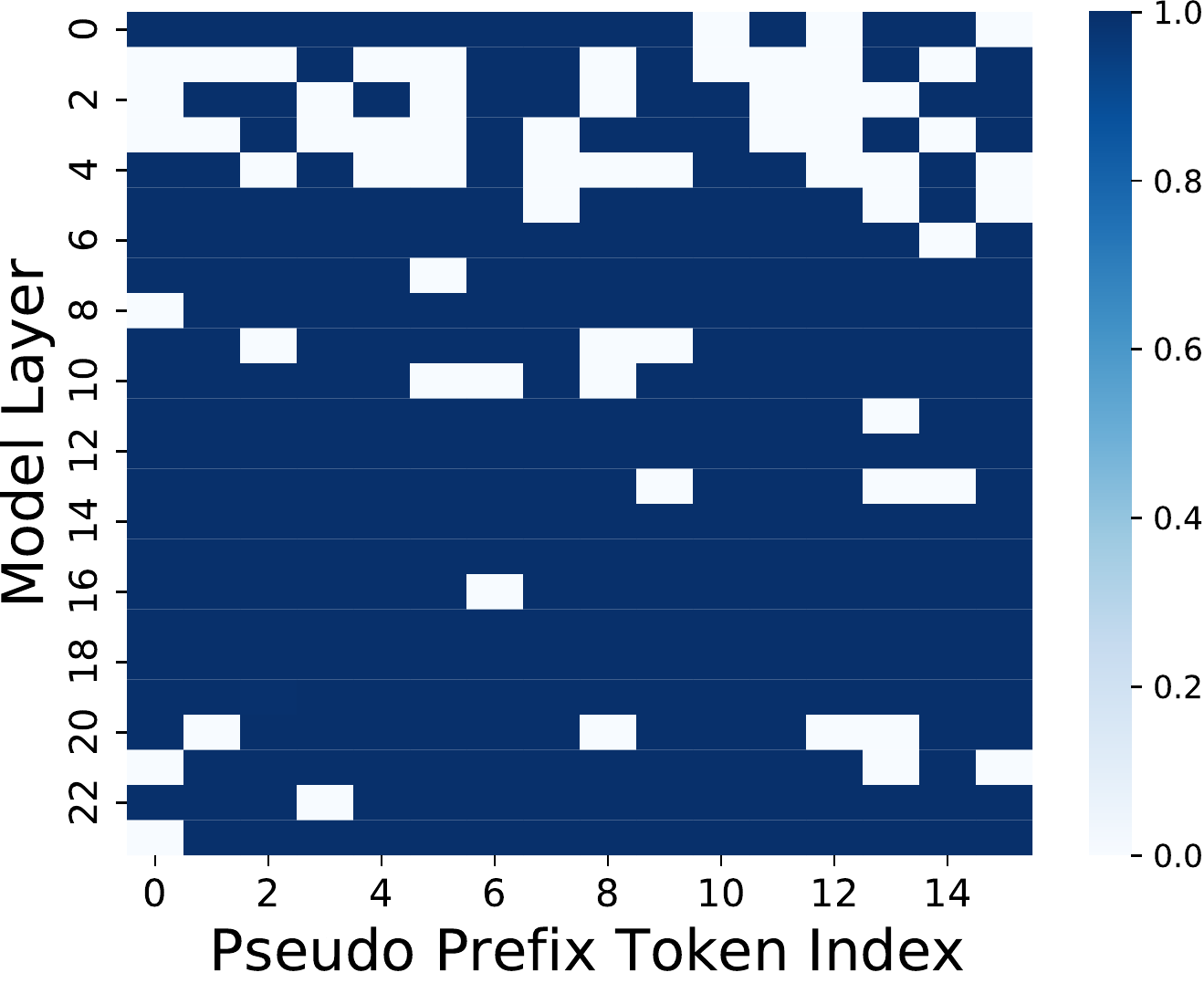}
    }
    \subfigure[CoNLL04]{
        \includegraphics[width=0.22 \textwidth]{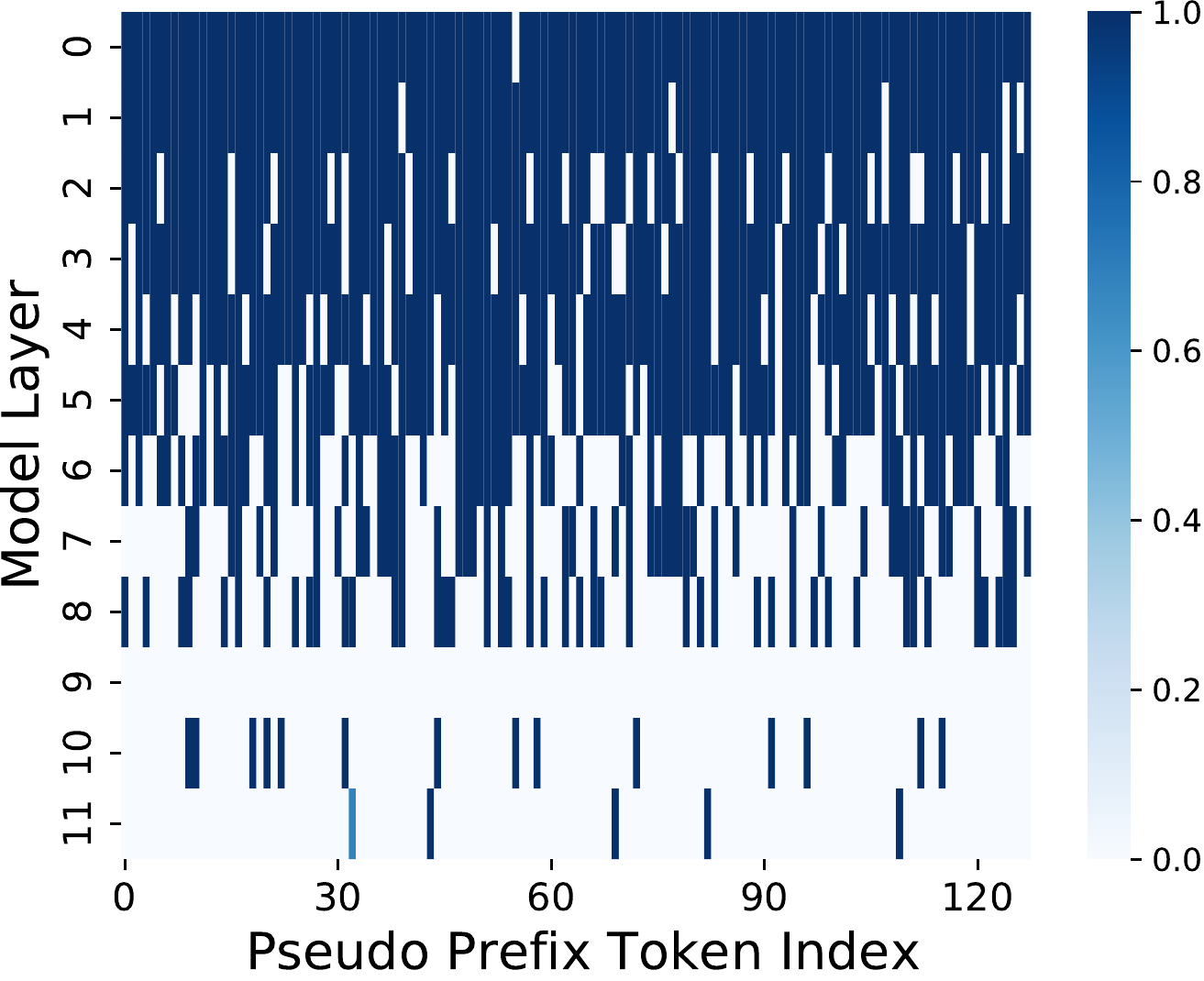}
    }
    \caption{Visualization of the learned weights of the prefix token for SuperGLUE task COPA on BERT-large and NER task CoNLL04 on BERT-base, with darker colors indicating higher weights.}
    \label{visualization}
\end{figure}

\subsection{Discussion}
\label{discussion1}
\paragraph{What is prefix weight distribution learned by APT?}
The gate mechanism for prefix serves as the key strategy of the proposed APT, where the learned prefix weight distribution turns out to be a critical point. Figure \ref{visualization} illustrates the gate weights of the pseudo prefix token for COPA and CoNLL04, respectively. It can be found that CoNLL04 is concerned with bottom layers in the language model which are regarded as phrase-level features, while COPA pays more attention to the higher layers, indicating semantic information. The observation is consistent with the characteristics of corresponding tasks. NER is a token-level task while COPA is a causal reasoning task sensitive to the semantics of sentences, which reminds us that it is worth placing various prefix tokens on specific layers according to the task properties.

\paragraph{Does \emph{variable} prefix work better than \emph{fixed} one?}
To verify the effectiveness of adaptive prefix under the proposed architecture, we wonder if the learned ratio at each layer can be directly transferred to P-Tuning v2. Taking the gate as a probing indicator, we reset the prefix length of P-Tuning v2 from fixed to variable in different layers based on the observation of the learned ratio (e.g. the distribution shown in Figure~\ref{visualization}). From the comparison between PT-2 and PT-2$^{*}$ in Table~\ref{discussion results}, we demonstrate that the variable prefix with less trainable parameters surprisingly outperforms the original implementation in fixed prefix. Nonetheless, it is also worth noting that there is still a gap between P-Tuning v2 with variable prefix and APT, where the latter continuously adjusts the weight of prefix during the training phase while the former only initializes with a one-time mask probing.

\paragraph{Whether the adaptive structure benefits the fine-tuning?}
Compared to P-Tuning v2, APT learns extra gated and scaled weights. To figure it out whether the improvement of APT is brought from more trainable parameters or the adaptive model structure, we adjust the hyper-parameter, i.e., enlarge the prefix length of P-Tuning v2 by 1.5 times to align the number of parameters with our APT. As shown in the comparison between PT-2$^{+}$ and APT of Table~\ref{discussion results}, we observe that APT still outperforms enlarged P-Tuning v2 with 1.9\%, 0.4\% on average for SuperGLUE and NER tasks respectively, validating the superiority of the gate mechanism.

\section{Conclusion}
In this paper, we investigate prefix tuning and assume that adaptive prefix is probably more efficient and effective than fixed prefix. Firstly, we propose APT that leverages the token-level and the layer-level gate mechanism which achieves an improvement of performance over original prefix tuning. Then, we illustrate the weight distribution learned by APT and take it as a probe, which validates the variable prefix can work better than the fixed one. The above experiments and analysis demonstrate that the adaptive prefix can be served as a promising strategy for parameter-efficient fine-tuning.

\section*{Limitations}
The proposed approach in this paper also suffers from certain limitations, i.e. we adapt APT on the encoder model and lack design for the other architectures such as decoder-only and encoder-decoder. In addition, it is better to generalize the key idea to other parameter-efficient learning approaches. A unified solution for existing work may be worth exploring in the future.

\bibliography{anthology,custom}
\bibliographystyle{acl_natbib}

\appendix
\section{Experimental Details}
\paragraph{Datasets}
In the full data setting, all train-dev-test splits follow P-Tuning v2 \cite{liu-etal-2022-p}. For low resources setting, to generate k-shot $(k=16, 32)$ datasets on SuperGLUE, the fixed set of random seed [11,21,42,87,100] is utilized to sample instances in training and development set, while the entire development set is treated as test set, where the average performance is reported in Table \ref{low resources results}.

\paragraph{Experimental Setting}
We grid search the learning rate over [5e-3, 7e-3, 1e-2, 1e-4], training epoch over [20, 40, 60, 80, 100, 120], batch size over [8, 16, 32], and random seeds over [11, 21, 42, 87, 100]. For a fair comparison, the prefix length utilized by APT is consistent with P-Tuning v2. In low resources setting, the batch size we used is 2. In Eq.(\ref{gated weight}), we take the hidden states of the first input token as representation in previous layer.

\begin{table*}[t]
\resizebox{1\textwidth}{!}{
\begin{tabular}{@{}llcccccccccc@{}}
\toprule
\multirow{2}{*}{\textbf{Model}} & \multirow{2}{*}{\textbf{Setting}} & \multicolumn{6}{c}{\textbf{SuperGLUE}} & \multicolumn{4}{c}{\textbf{NER}} \\ \cmidrule(lr){3-8} \cmidrule(lr){9-12} 
 &  & BoolQ & COPA & RTE & WiC & WSC & Avg. & CoNLL03 & CoNLL04 & OntoNotes & Avg. \\  \midrule
\multirow{4}{*}{BERT-large} & APT & \textbf{76.0} & \textbf{79.0} & \textbf{79.4} & \textbf{75.1} & \textbf{70.2} & \textbf{75.9} & 90.7 & \textbf{85.8} & \textbf{88.6} & \textbf{88.4} \\ \cmidrule(l){2-12} 
 & w/o token-level $\bm{\alpha}$ & 75.8 & 77.0 & 77.3 & 74.8 & 68.3 & 74.6 & 91.1 & 84.4 & 88.5 & 88.0 \\
 & w/o layer-level $\lambda$ & 75.4 & 74.0 & 76.9 & 74.6 & 68.3 & 73.8 & 90.7 & 83.7 & 88.4 & 87.6 \\
 & w/o hidden states $h$ & 74.7 & 76.0 & 75.8 & 74.6 & 68.3 & 73.9 & \textbf{91.2} & 84.0 & 88.6 & 87.9 \\
 \midrule
\multirow{4}{*}{RoBERTa-large} & APT & \textbf{84.8} & \textbf{94.0} & \textbf{89.9} & \textbf{74.6} & \textbf{68.3} & \textbf{82.3} & \textbf{92.7} & \textbf{89.0} & \textbf{89.8} & \textbf{90.5} \\ \cmidrule(l){2-12}
 & w/o   token-level $\bm{\alpha}$ & 84.3 & 88.0 & 88.1 & 73.0 & 65.4 & 79.8 & 92.2 & 88.7 & 89.5 & 90.1 \\
& w/o   layer-level $\lambda$ & 84.7 & 88.0 & 86.3 & 72.1 & 64.4 & 79.1 & 92.0 & 88.7 & 89.8 & 90.2 \\
 & w/o   hidden states $h$ & 83.9 & 91.0 & 87.0 & 72.9 & 64.4 & 79.8 & 92.2 & 88.7 & 89.4 & 90.1 \\ \bottomrule
\end{tabular}}
\caption{Ablation experiments on BERT-large and RoBERTa-large for two different level gate mechanisms and the hidden states from the previous layer. \textbf{bold}: the best score.}
\label{ablation study for large}
\end{table*}

\paragraph{Experimental Computation}
We use the pre-trained model BERT-base with 110M parameters, BERT-large with 335M parameters, RoBERTa-large with 355M parameters and DeBERTa-xlarge with 750M parameters. We conduct experiments on NVIDIA V100 or A100 GPUs for each task.

\section{Further Ablation Results}
We demonstrate further ablation results on BERT-large and RoBERTa-large as shown in Table \ref{ablation study for large}. It can be found that the beneficial impact of the three strategies and the observation is consistent with BERT-base in Section \ref{Ablation Study Section} in general. 

\section{Prefix Length}
\begin{figure}[t]
    \centering
    \subfigure[COPA]{
        \includegraphics[width=0.22 \textwidth]{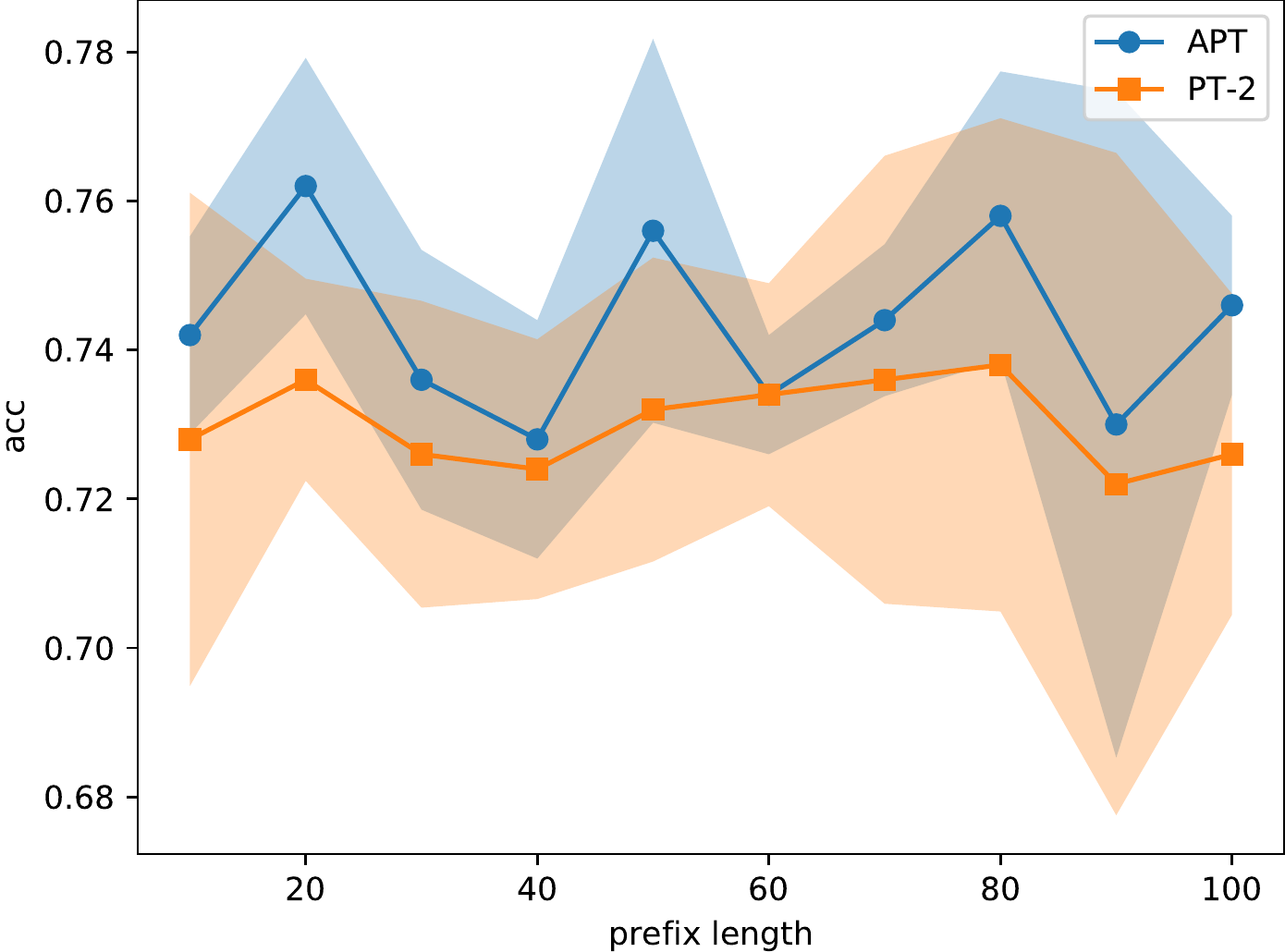}
    }
    \subfigure[WSC]{
        \includegraphics[width=0.22 \textwidth]{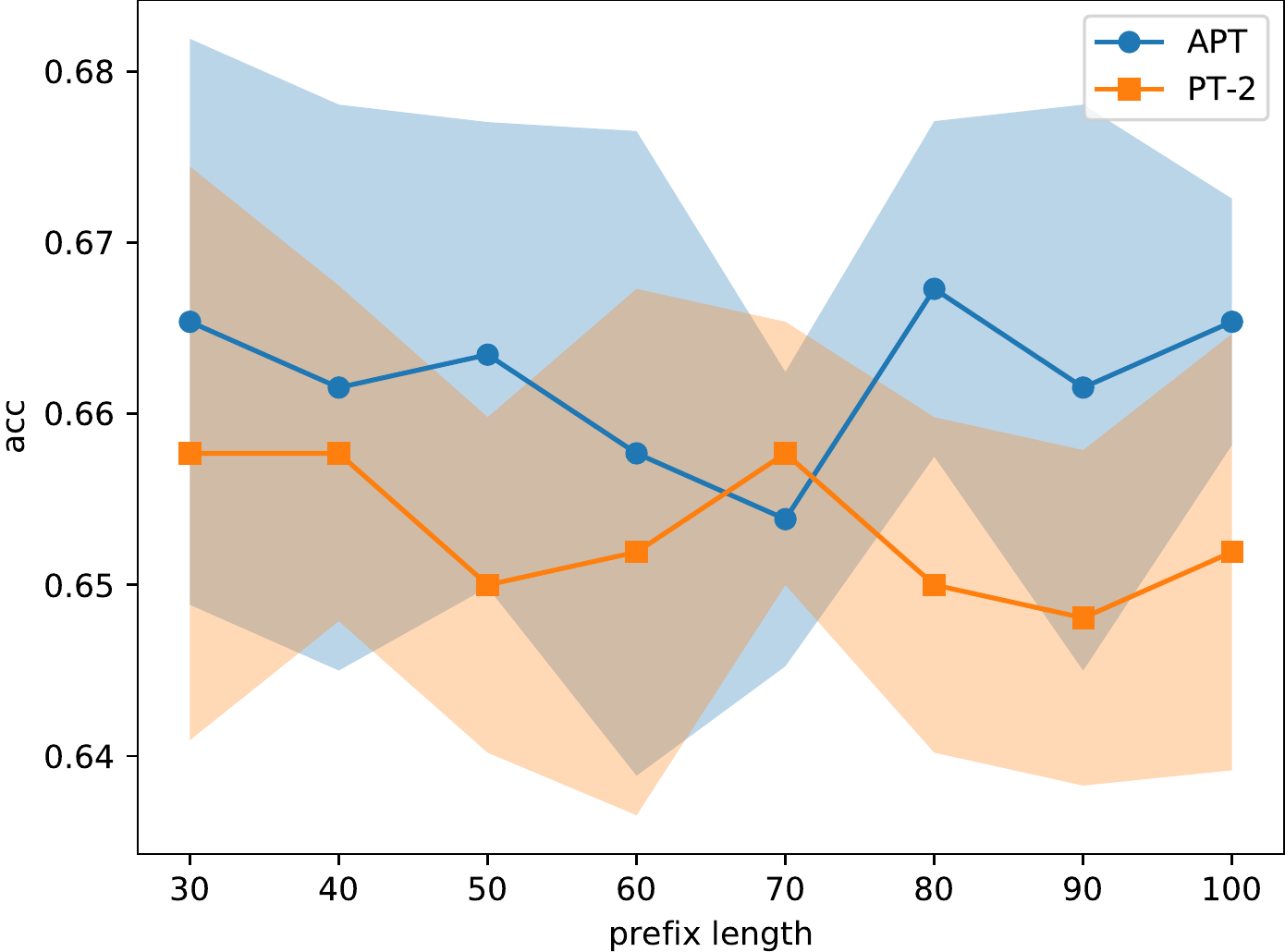}
    }
    \caption{The performance of APT and PT-2 on COPA and WSC in a range of prefix length on BERT-large.}
    \label{prefix length}
\end{figure}

The prefix length is an important hyper-parameter for prefix tuning and APT. Figure \ref{prefix length} illustrates the performance of APT and P-Tuning v2 with different prefix lengths over a range. It can be observed that APT is superior to P-Tuning v2 in most prefix length settings, indicating that APT has a relatively wider range of prefix length to achieve better performance.

\section{Scientific Artifacts}
We use datasets involving SuperGLUE \cite{NEURIPS2019_4496bf24} benchmark including BoolQ \cite{clark-etal-2019-boolq}, COPA \cite{roemmele2011choice}, RTE \cite{wang-etal-2018-glue}, WiC \cite{pilehvar-camacho-collados-2019-wic} and WSC \cite{levesque2012winograd} as well as 3 Named Entity Recognition (NER) tasks including CoNLL03 \cite{tjong-kim-sang-de-meulder-2003-introduction}, CoNLL04 \cite{carreras-marquez-2004-introduction}, and OntoNotes 5.0 \cite{AB2/MKJJ2R_2013}. The pre-trained model we used are BERT-base / large \cite{devlin-etal-2019-bert}, RoBERTa-large \cite{liu2019roberta} and DeBERTa-xlarge \cite{he2020deberta}. We use HuggingFace Transformers \cite{wolf-etal-2020-transformers} and P-Tuning v2 \cite{liu-etal-2022-p} as the codebase implemented by PyTorch \footnote{https://pytorch.org/}. They are all open-source and we only use for academic research which is consistent with their intended use.

\end{document}